\documentclass[10pt]{article} 
\usepackage[preprint]{tmlr}
\usepackage[utf8]{inputenc} 
\usepackage[T1]{fontenc} 
\usepackage{lmodern}

\usepackage{amsmath,amssymb,amsfonts,amsthm}
\usepackage{bm}
\usepackage{latexsym}
\usepackage{dsfont}
\usepackage{mathrsfs}

\usepackage{graphicx}
\usepackage{subcaption}
\usepackage{afterpage}
\usepackage{placeins}

\usepackage{tikz}
  \tikzset{>=stealth}
  \usetikzlibrary{trees,arrows,positioning,chains,external}
  \usetikzlibrary{positioning,shadows.blur,arrows.meta,shapes.multipart}

\usepackage[ruled,vlined]{algorithm2e}

\usepackage{url}
\usepackage{hyperref}
\usepackage{cleveref}
\usepackage{verbatim}
\usepackage{lscape}
\usepackage{rotating}
\usepackage{adjustbox}
\usepackage{booktabs}
\usepackage{array}
\usepackage{arydshln}
\usepackage{siunitx}
\usepackage{xcolor}

\newcommand{\dd}{\mathrm{d}}
\newcommand{\bh}{\mathbf{h}}
\newcommand{\bx}{\mathbf{x}}

\title{
SDE-Attention: Latent Attention in SDE-RNNs for Irregularly Sampled Time Series with Missing Data\\
}

\author{%
\name Yuting Fang \email z5518340@ad.unsw.edu.au \\
\addr School of Mathematics and Statistics\\
UNSW Sydney
\AND
\name Quoc Le Gia \email q.legia@unsw.edu.au \\
\addr School of Mathematics and Statistics\\
UNSW Sydney
\AND
\name Flora D. Salim \email flora.salim@unsw.edu.au \\
\addr School of Computer Science and Engineering\\
UNSW Sydney
}
\begin{document}

\maketitle

\begin{abstract}
Irregularly sampled time series with substantial missing observations are common in healthcare and sensor networks. We introduce SDE--Attention, a family of SDE--RNNs equipped with channel-level attention on the latent pre–RNN state, including channel recalibration, time-varying feature attention, and pyramidal multi-scale self-attention. We therefore conduct a comparison on a synthetic periodic dataset and real-world benchmarks, under varying missing rate. Latent-space attention consistently improves over a vanilla SDE--RNN. On the univariate UCR datasets, the LSTM-based time-varying feature model SDE--TVF-L achieves the highest average accuracy, raising mean performance by approximately 4, 6, and 10 percentage points over the baseline at 30\%, 60\% and 90\% missingness, respectively (averaged across datasets). On multivariate UEA benchmarks, attention-augmented models again outperform the backbone, with SDE--TVF-L yielding up to around 7\% gain in mean accuracy under high missingness. Among the proposed mechanisms, time-varying feature attention is the most robust on univariate datasets. In multivariate datasets, different types of attention excel in different tasks, showing that SDE--Attention can be flexibly adapted to the structure of each problem \url{https://anonymous.4open.science/r/SDE-Attention-BEFB}
\end{abstract}

\section{Introduction}

Irregularly sampled time series with substantial missing observations are ubiquitous in healthcare, environmental monitoring, and sensor networks\citep{Seqlink2024}. For instance, Electronic health records contain measurements such as lab tests and vital signs that are collected at patient specific times and often sparsely observed\citep{shukla2018modeling}. Traditional time series models, including Recurrent Neural Network (RNN) \citep{rumelhart1986learning}, Long Short-term Memory (LSTM)\citep{graves2012long} and Gated Recurrent Unit (GRU) \citep{chung2014empirical}, treat observed samples as consecutive discrete sequences, struggling with irregularly-sampled or partially-observed data \citep{mozer2017discrete}. GRU-D\citep{che2016grud} explicitly encodes missingness patterns through decay mechanisms and masking, however, it still operates in discrete time and does not leverage the continuous nature of the latent dynamics.

Recently, a powerful alternative for handling irregular dataset, the sequential deep time series model, has gradually gained prominence. Neural ordinary differential equations (Neural ODE) parameterize the derivative of a hidden state and use an ODE solver to obtain continuous transformations \citep{chen2018neural}. Latent ODEs and ODE--RNNs extend this idea to irregularly sampled time series by combining continuous time latent dynamics with event driven updates \citep{rubanova2019latent}. Neural controlled differential equations (Neural CDEs) further tuning time series modeling in rough path theory, providing a principled framework for partially observed, irregular multivariate sequences \citep{kidger2020neuralcde}. Furthermore, to catch the uncertainty inside the continuous trajectory, \cite{liu2019neural} add the diffusion term to the Neural ODE, obtaining the Neural Stochastic differential equation, which can both catch the certain trajectory and the randomness of latent dynamics. Additionally, as the extension of ODE-based models, SDE--RNN\citep{dahale2023general} is also designed to solve irregularly sampled time series in the same way, with additive randomness. These works demonstrate that stochastic continuous time latent dynamics can naturally handle irregular sampling and variable length trajectories. 

Despite these advances, important questions remain about the robustness of neural stochastic differential equation models, particularly under severe missingness \citep{oh2024stable}. While SDE-based architectures provide a principled way to capture both continuous trajectories and latent randomness, they typically treat all channels in hidden space uniformly and rely on the SDE--RNN backbone to implicitly infer which features and time points are informative. This limits their ability to explicitly disentangle the contribution of different variables and to remain stable when the observation pattern becomes highly sparse or irregular.

Concurrently, a substantial body of research has demonstrated that channel-level attention can significantly enhance the robustness and interpretability of discrete time models. Channel-wise recalibration mechanisms such as squeeze-and-excitation (SE) blocks adaptively rescale feature maps based on global statistics \citep{hu2018senet}, and time corrected residual attention networks (TCRAN) propose specialized channel attention blocks for multivariate time series classification \citep{zhu2022tcran}. For long sequences, pyramidal and multi-scale attention architectures, such as Pyraformer\citep{liu2022pyraformer}, capture long range dependencies at reduced computational cost, while Transformer variants for time series exploit self-attention to model complex temporal and cross-channel interactions \citep{zerveas2021tstransformer}. However, these attention mechanisms have been studied almost exclusively in discrete-time architectures. It therefore remains unclear how to best incorporate channel-level attention into SDE--RNNs: which types of attention are most effective for SDE-based models, where attention should be inserted at latent space, and how these design choices interact with severe observation sparsity. 

In this work, we address these questions by proposing SDE-Attention, a unified framework for augmenting SDE--RNNs with channel-level attention time series. We keep a continuous-time SDE--RNN backbone fixed and attach three plug-and-play attention modules at the latent level: (i) a batch-aggregated latent channel attention module that gates latent states in a time specific but trajectory-shared fashion; (ii) a time-varying feature attention module instantiated with either an LSTM or Transformer encode; and (iii) a multi-scale pyramidal self-attention module that aggregates information across multiple temporal resolutions before feeding the hidden state into the GRU. This design disentangles where attention is applied from how it models channel importance, allowing us to systematically evaluate their impact within a controlled SDE--RNN setting.

Overall, our contributions are threefold: (i) we introduce SDE-Attention, a family of attention-augmented SDE--RNN architectures; (ii) we provide, to the best of our knowledge, the first systematic study of channel-level attention in SDE-based sequence models under extreme observation sparsity; (iii) We provide empirical guidance on when to prefer simple channel recalibration vs. more expressive TVF modules across univariate and multivariate benchmarks under different missingness levels.

\section{Preliminaries}
\label{sec:Preliminaries}
\subsection{Irregular time series}

We consider multivariate time series observed at irregular time stamps.
For each sequence we write
\(
\{(x_i, t_i)\}_{i=1}^N
\),
where \(x_i \in \mathbb{R}^D\) denotes a \(D\)-dimensional observation at
time \(t_i\), with
\(0 \le t_1 < \cdots < t_N\).
Missingness is represented by a binary mask
\(m_i \in \{0,1\}^D\) and the actually observed values are
\(\tilde x_i = m_i \odot x_i\).
Given such irregular and partially observed sequences, the goal is to learn
a model that maps each trajectory to a label \(y\) (time series classification)
while remaining robust to severe observation sparsity.
\subsection{Models}
\textbf{Neural ODEs}\citep{chen2018neural} has the form:
\begin{equation}\
\mathbf{h}(0) = \mathbf{h}_0  \quad \frac{dh}{dt} = f_\theta(t,\mathbf{h}(t)) \label{neuralODEs},
\end{equation}
where $\mathbf{h}_0 \in \mathbb{R}^{d_1*\dots *d_k}$ is a any-dimensional tensor, $\theta$ represent the vector of parameters of $f$, which is a  neural network.

\textbf{Neural SDEs}\citep{liu2019neural} extend Neural ODEs by
introducing randomness in the dynamics:
\begin{equation}
  \dd \bh_t = f_\theta(\bh_t, t)\,\dd t + g_\theta(\bh_t, t)\,\dd \mathbf{W}_t,
  \label{eq:neural-sde}
\end{equation}
where $\mathbf{W}_t$ is a Brownian motion; $f_\theta$ parameterize the drift network; $g_\theta$ is the diffusion network.

\textbf{SDE-RNN}\citep{dahale2023general} When combined with RNN updates at observation times, Neural ODEs yield ODE–RNNs for irregular time series\citep{rubanova2019latent}. SDE–RNN extends this idea by replacing the deterministic ODE with a Neural SDE, using numerical schemes such as Euler–Maruyama\citep{higham2001algorithmic} along a fixed Brownian path. And the Algorithm \ref{alg:sde_rnn} shows the rough structure of this.

\begin{algorithm}[H]
\caption{The SDE\textendash RNN. }
\KwIn{Data points and timestamps $\{(x_i, t_i)\}_{i=1..N}$; SDE components $f_\theta, g_\theta$; SDE solver (e.g., \texttt{sdeint}); Brownian generator $\mathcal{B}$ (path/interval)}
$h_0 \gets \mathbf{0}$\;
\For{$i \in \{1,2,\ldots,N\}$}{
  {\color{blue}$h'_i \gets \operatorname{SDEsolve}\!\big(f_\theta, g_\theta,\, h_{i-1},\, (t_{i-1}, t_i),\, \mathcal{B}\big)$}\tcp*{Integrate $\mathrm{d}h=f\,\mathrm{d}t+g\,\mathrm{d}W$ to reach $t_i$}
  $h_i \gets \operatorname{RNNCell}\!\big(h'_i,\, x_i\big)$\tcp*{Assimilate the observation at $t_i$}
}
$o_i \gets \operatorname{OutputNN}(h_i)$ \; for all $i=1..N$\;
\Return{$\{o_i\}_{i=1..N};\; h_N$}
\label{alg:sde_rnn}
\end{algorithm}

\subsection{channel-level attention mechanisms for time series}
According to the properties of time series, there are several types of attention, including channel, temporal and multi-scale attention, can work on it. In this article, we will focus on the channel level attention mechanism for latent state. The main idea of channel Attention is to recalibrate feature dimensions by learning data-dependent weights per channel, also can vary through time.

\section{SDE-RNN with Attention}
\label{sec:method}

\subsection{SDE-RNN backbone}
\label{subsec:sde-rnn-backbone}
\begin{figure}
    \centering    
    \includegraphics[width=0.3\linewidth]{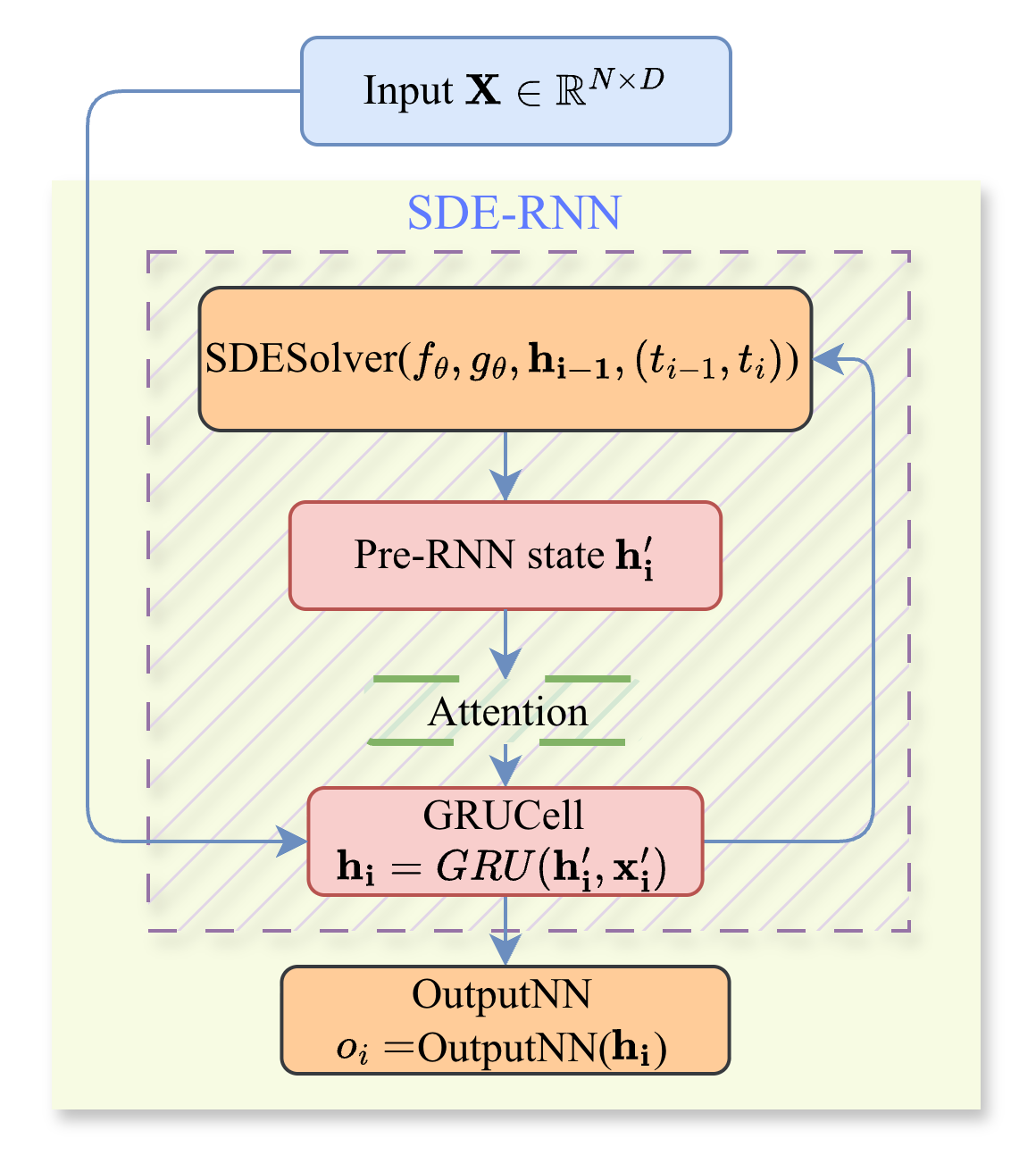}
    \caption{SDE-RNN backbone. Between observation times, the latent state is evolved by an SDE solver; at each observation time $t_i$, a GRU cell updates the state using the reweighted latent $h'_i$. The green bracket indicates the insertion point of latent channel attention; for other attention variants, the same backbone is used with a different attention module at the corresponding latent location.}
    \label{fig:sde-rnn=attn}
\end{figure}
We adopt an SDE--RNN backbone as illustrated in
Fig.~\ref{fig:sde-rnn=attn}.  
Given an irregular time series
$\{(x_i, t_i)\}_{i=1}^N$ and the post-RNN hidden state $ \bh_{i-1}$ at the
previous observation time $t_{i-1}$, we first evolve the latent state between
$t_{i-1}$ and $t_i$ by integrating the Neural SDE in
Eq.~\eqref{eq:neural-sde} along a fixed Brownian path:
\begin{equation}
  \bh'_i = \mathrm{SDESolver}\bigl(f_\theta, g_\theta, \bh_{i-1}, (t_{i-1}, t_i)\bigr),
\end{equation}
where $\bh'_i$ denotes the \emph{pre-RNN} state at time $t_i$.

An attention module then optionally transforms the pre-RNN state, $\tilde{\bh}'_i$.  The RNN update at $t_i$ is performed by a GRU cell:
\begin{equation}
  \bh_i = \mathrm{GRUCell}\bigl(\tilde{\bh}'_i, \bx'_i\bigr),
\end{equation}
where $\bh_i$ is the post-RNN hidden state.  At each step, a prediction
is produced from the pre-RNN state via a small output network
\begin{equation}
  o_i = \mathrm{OutputNN}(\bh'_i).
\end{equation}

\subsection{SDE-Attention framework}
\label{subsec:sde-attention-framework}
\begin{figure*}[!ht]
  \centering
  \begin{subfigure}[b]{0.3\textwidth}
    \centering
    \includegraphics[width=\textwidth]{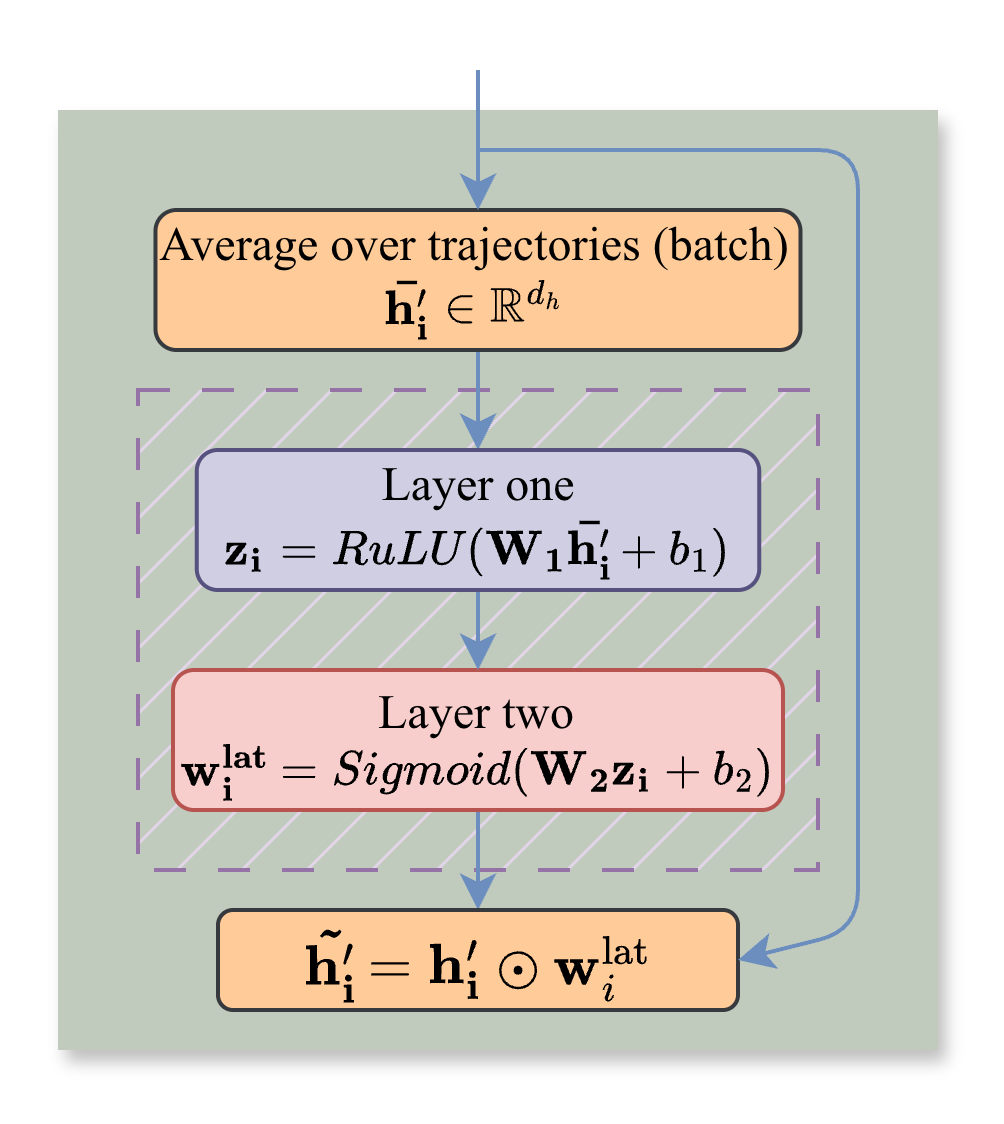}
    \caption{Static channel attention}
    \label{fig:latentchannel}
  \end{subfigure}
  \hfill
  \begin{subfigure}[b]{0.3\textwidth}
    \centering
    \includegraphics[width=\textwidth]{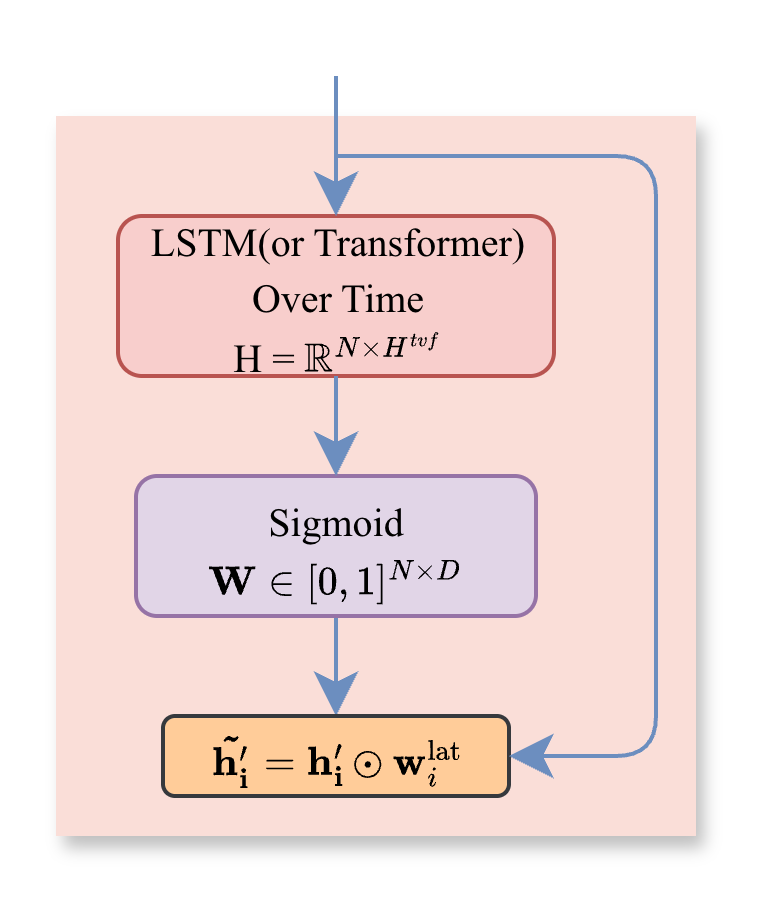}
    \caption{TVF attention}
    \label{fig:tvf}
  \end{subfigure}
  \hfill
  \begin{subfigure}[b]{0.3\textwidth}
    \centering
    \includegraphics[width=\textwidth]{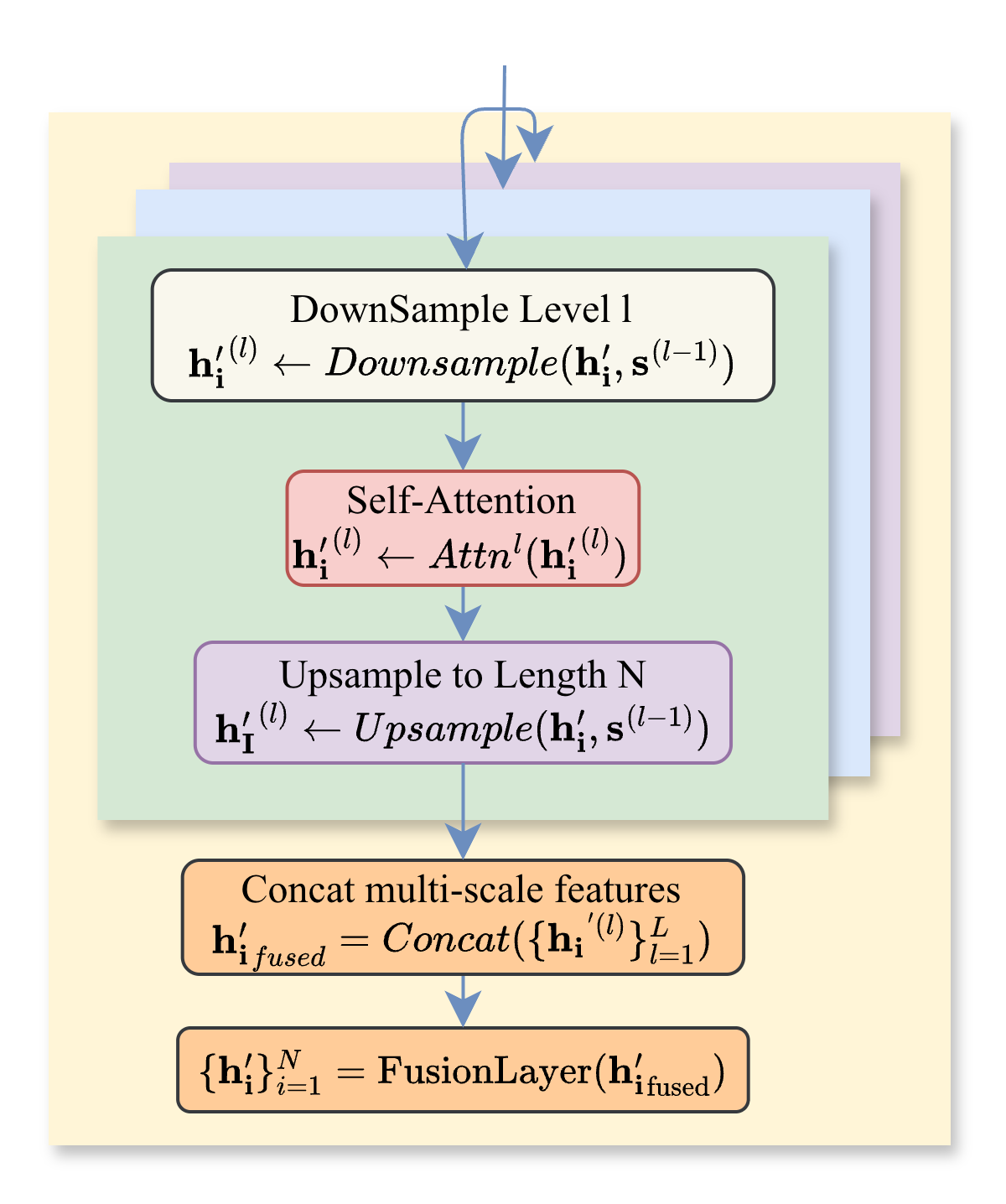}
    \caption{Pyramidal attention}
    \label{fig:pyramidal}
  \end{subfigure}

  \caption{
  Attention modules used in our SDE-Attention framework.
  (a) Static channel attention reweights latent dimensions based on a global summary of hidden states.
  (b) TVF uses an LSTM or Transformer along time to produce time-varying feature-wise gates.
  (c) The pyramidal module builds multi-scale representations via repeated downsampling, self-attention, and upsampling, and fuses features from all levels.
  }
  \label{fig:attn-modules}
\end{figure*}

\subsubsection{Static Channel Attention}

Latent channel attention reweights each latent dimension using batch-level statistics at every time step (Figure~\ref{fig:latentchannel}). For a fixed time index $i$, we aggregate all trajectories in the mini-batch into a summary vector and pass it through a small two-layer MLP to obtain a gate vector for that time step. This gate is shared across the batch and applied element-wise to all pre-RNN states at time $i$.

We denote this operation by
\[
\tilde{\mathbf{h}}'_i = \mathcal{A}_{\text{lat}}\bigl(\{\mathbf{h}'_{i,b}\}_{b=1}^B\bigr),
\]
where $\{\mathbf{h}'_{i,b}\}$ are pre-RNN states in the batch and
$\tilde{\mathbf{h}}'_i$ is the gated latent state fed into the GRUCell.
Intuitively, channels that are consistently informative across trajectories at time $i$
receive larger weights, while noisy channels are downweighted.


\subsubsection{Time-Varying Feature Attention}

Time-varying feature attention (\textsc{TVF}, Figure~\ref{fig:tvf}) lifts channel
attention to the fully time dependent setting. Instead of a single gate per
feature, TVF assigns a distinct weight to each feature at each time step.

Given an sequence $\mathbf{h'_i}$, a temporal encoder
$\Psi_{\text{attn}}$ (either a bi-LSTM or a Transformer) produces
context-aware hidden states for all time steps. These hidden states are then
mapped to a matrix of gates $W \in [0,1]^{N \times D}$ via a linear layer with
sigmoid activation. The final attended sequence is obtained by
element-wise reweighting,
\[
\tilde{\mathbf{h'_i}} = \mathcal{A}_{\text{tvf}}(\mathbf{h'_i}) = \mathbf{h'_i} \odot W,
\]
where $\mathcal{A}_{\text{tvf}}$ denotes the TVF module. In contrast to TVF allows channel importance to change over time, which is useful when different sensors become relevant at different phases of a trajectory, at the cost of higher computational complexity.


\subsubsection{Pyramidal Attention}

Pyramidal attention (Figure~\ref{fig:pyramidal})
builds a multi-scale representation of the hidden dynamics before it is fed into the GRU. Instead of running self-attention only at the original
resolution, it repeatedly downsamples the sequence to obtain coarser views, applies self-attention at each scale, then upsamples and fuses all scales
into a single attended sequence.

Concretely, given $\mathbf{h'_i}$, the module constructs
$L$ levels indexed by $l = 0,\ldots,L-1$. Each level uses a stride
$s_l = \texttt{stride\_base}^{\,l}$ (in our implementation
$\texttt{stride\_base}=2$, so the strides are $1,2,4,\ldots$). The
\emph{downsample} operation simply keeps every $s_l$-th time step:
fine levels (small $s_l$) see the sequence almost unchanged, while coarse
levels (large $s_l$) see a much shorter sequence that emphasises long-range
structure.

At each level $l$, an attention block operates on the downsampled sequence and produces a transformed sequence at that scale. The result is then
\emph{upsampled} back to length $N$ using 1D linear interpolation along the
time axis, so that all levels are temporally aligned. In other words,
each level produces a sequence in $\mathbb{R}^{N \times D}$ that encodes
patterns at its own temporal resolution.

Finally, the $L$ upsampled sequences are concatenated along the feature
dimension and passed through a linear fusion layer, yielding an attended
sequence 
\[
\tilde{\mathbf{h'_i}} = \mathcal{A}_{\text{pyr}}(\mathbf{h'_i})
\] 
with the same shape as the input. Intuitively, pyramidal attention allows the model to combine fine-grained
local information (from low levels) with coarse, long-range trends (from
high levels) in a single representation.

\section{Experimental Setup}
\label{sec:experiments-setup}

\subsection{Datasets and experiments setting}
We evaluate our models on uni-variate and multivariate time series datasets including a generated Periodic time series, UEA and UCR.

\textbf{Periodic} We generate toy dataset of 1,000 periodic trajectories with temporally correlated noise to evaluate model robustness. Each trajectory is defined as:
\begin{equation}
y(t) = A(t) \cdot \sin(\phi(t)) + z_0 + \eta(t)
\end{equation}
where $A(t)$ is the time-varying amplitude and $\phi(t) = \int 2\pi f(t) dt$ is the integrated phase from frequency $f(t)$. Each trajectory has 100 irregularly-sampled time points.
we model measurement noise $\eta(t)$ using an Ornstein-Uhlenbeck (OU) process, providing a more realistic representation of measurement noise in physical and biological systems where consecutive observations are temporally dependent. We train all models for 500 iterations and systematically vary observation rates at $\{10\%, 20\%, 30\%, 40\%\}$. Other settings are just same to the UCR and UEA.

\textbf{UCR/UEA}(\cite{dau19ucr}, \cite{bagnall2018uea}): They are the University of California Riverside (UCR) and University of East Anglia (UEA) Time Series Classification Repository, covering univariate and multivariate time series datasets from various real-world applications. We test on around 18 datasets, from both UEA and UCR(~\ref{tab:dataset-classification-5cat}). All models are trained for 100 iterations, using SDE-RNN as the base architecture with 50 units single layer MLP in the drift and diffusion networks, across four missing data scenarios (0\%, 30\%, 60\%, 90\% missing) to study robustness under missingness. For each missingness level, we independently drop each observed value with certain probability, yielding an MCAR mask applied per time step and per channel. 

\begin{table*}[!ht]
\centering
\caption{Fine-grained domain classification of 18 selected datasets}
\label{tab:dataset-classification-5cat}
\small
\begin{adjustbox}{max width=\linewidth}
\begin{tabular}{lllrrrrl}
\multicolumn{8}{l}{\footnotesize N=Samples, C=Classes, D=Dimensions, L=Length}\\
\toprule
\textbf{Category} & \textbf{Subdomain} & \textbf{Dataset} & \textbf{N} & \textbf{C} & \textbf{D} & \textbf{L} & \textbf{Source} \\
\midrule
\multicolumn{8}{l}{\textbf{1. Healthcare \& Biomedical}} \\
\midrule
Neuroscience & Clinical Neurology & Epilepsy & 275 & 4 & 3 & 206 & UEA \\
Neuroscience & Brain-Computer Interface & FingerMovements & 316 & 2 & 28 & 50 & UEA \\
Neuroscience & Neural Self-Regulation & SelfRegulationSCP2 & 380 & 2 & 7 & 1152 & UEA \\
Medical Imaging & Bone Morphology & MiddlePhalanxOutlineAgeGroup & 554 & 3 & 1 & 80 & UCR \\
Medical Imaging & Bone Morphology & ProximalPhalanxOutlineAgeGroup & 605 & 3 & 1 & 80 & UCR \\
Medical Imaging & Bone Morphology & ProximalPhalanxOutlineCorrect & 891 & 2 & 1 & 80 & UCR \\
Medical Imaging & Bone Morphology & ProximalPhalanxTW & 605 & 6 & 1 & 80 & UCR \\
\midrule
\multicolumn{8}{l}{\textbf{2. Human Activity \& Interaction}} \\
\midrule
HCI & Gesture Recognition & UWaveGestureLibrary & 120 & 8 & 3 & 315 & UEA \\
Movement Science & Activity Recognition & BasicMotions & 80 & 4 & 6 & 100 & UEA \\
Language & Sign Language Recognition & Libras & 360 & 15 & 2 & 45 & UEA \\
Language & Speech Articulation & ArticularyWordRecognition & 575 & 25 & 9 & 144 & UEA \\
\midrule
\multicolumn{8}{l}{\textbf{3. Industrial \& Engineering}} \\
\midrule
Manufacturing & Semiconductor QC & Wafer & 7164 & 2 & 1 & 152 & UCR \\
Manufacturing & Sensor Monitoring & MoteStrain & 1272 & 2 & 1 & 84 & UCR \\
Robotics & Surface Detection & SonyAIBORobotSurface2 & 980 & 2 & 1 & 65 & UCR \\
Vision & Shape Recognition & ERing & 300 & 6 & 4 & 65 & UEA \\
\midrule
\multicolumn{8}{l}{\textbf{4. Natural Sciences}} \\
\midrule
Geophysics & Seismology & Earthquakes & 461 & 2 & 1 & 512 & UCR \\
Agriculture & Quality Assessment & Strawberry & 983 & 2 & 1 & 235 & UCR \\
\midrule
\multicolumn{8}{l}{\textbf{5. Synthetic Benchmark}} \\
\midrule
Synthetic & Pattern Recognition & TwoPatterns & 5000 & 4 & 1 & 128 & UCR \\
\bottomrule
\end{tabular}
\end{adjustbox}
\end{table*}
We train all models with the Adam optimiser and report test accuracy averaged over multiple random seeds. We also report mean and standard deviation across datasets to summaries overall performance.
\subsection{Models and baselines}
We test four different types of attention to the SDE-RNN, where vanilla \textbf{SDE-RNN}\cite{dahale2023general} is the baseline. Therefore, we have the following models:
\textbf{SDE-StaticChannel}(SDE-SCHA): SDE-RNN with Latent channel attention, which is the attention working on the latent space.
\textbf{SDE-TVF-LSTM}(SDE-TVF-L): SDE-RNN with time-varying feature attention using an LSTM encoder and also reweighs the hidden data.
\textbf{SDE-TVF-Transformer}(SDE-TVF-T): Same with SDE-TVF-L, but uses a Transformer encoder.
\textbf{SDE-Pyramidal}(SDE-PYR): SDE-RNN with pyramidal multi-scale attention, still work on the hidden space.

\section{Results}
\label{sec:results}
\subsection{Performance under fully observed data}

\begin{table}[!ht]
\centering
\scriptsize
\setlength{\tabcolsep}{4pt}
\renewcommand{\arraystretch}{1.1}
\caption{Classification accuracy (mean (std)) on selected UCR datasets without missing observations. Best result per dataset is in bold.}
\label{tab:ucr-miss-00-selected-new}
\begin{adjustbox}{max width=\linewidth}
\begin{tabular}{lrrrrr}
\toprule
Dataset & \multicolumn{5}{c}{Accuracy} \\
\cmidrule(lr){2-6}
 & SDE-RNN & SDE-PYR & SDE-TVF-T & SDE-SCHA & SDE-TVF-L \\
\midrule
MiddlePhalanxOutlineAgeGroup   & 0.604 (0.009) & \textbf{0.612 (0.012)} & 0.594 (0.004) & 0.599 (0.008) & 0.607 (0.003) \\
Earthquakes                    & 0.572 (0.125) & 0.758 (0.009) & 0.681 (0.000) & \textbf{0.776 (0.007)} & 0.751 (0.004) \\
MoteStrain                     & 0.758 (0.002) & 0.765 (0.008) & 0.774 (0.006) & 0.773 (0.006) & \textbf{0.777 (0.002)} \\
ProximalPhalanxOutlineAgeGroup & 0.848 (0.012) & \textbf{0.866 (0.002)} & 0.859 (0.004) & 0.856 (0.006) & 0.647 (0.300) \\
ProximalPhalanxOutlineCorrect  & 0.708 (0.008) & 0.695 (0.012) & 0.696 (0.025) & 0.744 (0.020) & \textbf{0.765 (0.003)} \\
ProximalPhalanxTW              & \textbf{0.781 (0.007)} & 0.777 (0.001) & 0.773 (0.007) & 0.774 (0.007) & 0.767 (0.003) \\
SonyAIBORobotSurface2          & 0.697 (0.015) & 0.706 (0.008) & \textbf{0.748 (0.033)} & 0.725 (0.018) & 0.743 (0.011) \\
Strawberry                     & 0.589 (0.035) & 0.605 (0.054) & 0.612 (0.044) & 0.711 (0.046) & \textbf{0.764 (0.013)} \\
TwoPatterns                    & 0.545 (0.051) & 0.471 (0.057) & 0.669 (0.233) & \textbf{0.716 (0.210)} & 0.502 (0.004) \\
Wafer                          & 0.892 (0.000) & 0.912 (0.028) & 0.934 (0.030) & 0.971 (0.015) & \textbf{0.987 (0.004)} \\
\bottomrule
\end{tabular}
\end{adjustbox}
\end{table}

\textbf{UCR} Table~\ref{tab:ucr-miss-00-selected-new} reports classification accuracy on ten UCR datasets without missing observations. Across most datasets, all attention-augmented variants of SDE–RNN (SDE–PYR, SDE–TVF–T, SDE–SCHA, SDE–TVF–L) match or improve upon the vanilla SDE–RNN baseline, indicating that latent-space attention is beneficial even when the data are fully observed. The gains are generally moderate rather than dramatic, but they form a consistent positive trend.

\subsection{Robustness to missing observations}

\textbf{Periodic Dataset} Table~\ref{tab:periodic-mse-interp} reports the interpolation MSE on the periodic toy dataset. Across all observation rates, attaching attention in the latent space consistently improves over the vanilla SDE–RNN baseline. When only 10\%of points are observed, all variants reduce MSE relative to SDE–RNN (0.583), with SDE–PYR achieving the lowest error (0.518). As the observation rate increases to 20\%–40\%, the gaps become smaller but remain systematic: both time-varying feature models (SDE–TVF–L/T) and the latent channel module (SDE–LC) outperform the baseline at every density. At 30\% and 40\% observed points, SDE–TVF–T attains the best performance (0.234 and 0.180, respectively), while SDE–LC is a close second (0.240 and 0.182).
\begin{table}[!ht]
\centering
\scriptsize
\setlength{\tabcolsep}{6pt}
\renewcommand{\arraystretch}{1.1}
\caption{Mean squared error on the periodic dataset (Interpolation only)}
\label{tab:periodic-mse-interp}
\begin{tabular}{lcccc}
\toprule
 & \multicolumn{4}{c}{Interpolation (\%Observed Points)} \\
\cmidrule(lr){2-5}
\textbf{Model} & \textbf{10} & \textbf{20} & \textbf{30} & \textbf{40} \\
\midrule
\textbf{SDE-RNN}   & 0.583(0.036) & 0.355(0.035) & 0.254(0.021) & 0.197(0.010) \\
\textbf{SDE-PYR}   & \textbf{0.518(0.027)} & 0.354(0.017) & 0.244(0.011) & 0.189(0.009) \\
\textbf{SDE-TVF-L} & 0.539(0.020) & \textbf{0.336(0.010)} & \textbf{0.234(0.013)} & 0.182(0.008) \\
\textbf{SDE-TVF-T} & 0.558(0.017) & 0.338(0.023) & 0.234(0.009) & \textbf{0.180(0.006)} \\
\textbf{SDE-SCHA}    & 0.553(0.020) & 0.339(0.016) & 0.240(0.005) & 0.182(0.006) \\
\bottomrule
\end{tabular}
\end{table}

\begin{table}[!ht]
\centering
\scriptsize
\setlength{\tabcolsep}{4pt}
\renewcommand{\arraystretch}{1.1}
\caption{Classification accuracy (mean (std)) on selected UCR datasets with 60\% missing observations. Best result per dataset is in bold.}
\label{tab:ucr-miss-60-selected-new}
\begin{adjustbox}{max width=\linewidth}
\begin{tabular}{lrrrrr}
\toprule
Dataset & \multicolumn{5}{c}{Accuracy} \\
\cmidrule(lr){2-6}
 & SDE-RNN & SDE-PYR & SDE-TVF-T & SDE-SCHA & SDE-TVF-L \\
\midrule
MiddlePhalanxOutlineAgeGroup   & 0.386 (0.088) & 0.422 (0.019) & \textbf{0.496 (0.034)} & 0.490 (0.035) & 0.488 (0.022) \\
Earthquakes                    & 0.640 (0.085) & 0.745 (0.000) & 0.591 (0.073) & 0.611 (0.081) & \textbf{0.753 (0.003)} \\
MoteStrain                     & 0.498 (0.019) & 0.505 (0.025) & 0.534 (0.052) & 0.616 (0.067) & \textbf{0.637 (0.071)} \\
ProximalPhalanxOutlineAgeGroup & 0.744 (0.036) & 0.712 (0.032) & \textbf{0.789 (0.011)} & 0.787 (0.026) & 0.767 (0.023) \\
ProximalPhalanxOutlineCorrect  & 0.647 (0.012) & 0.619 (0.031) & \textbf{0.680 (0.007)} & 0.673 (0.012) & 0.637 (0.005) \\
ProximalPhalanxTW              & 0.657 (0.048) & 0.614 (0.018) & 0.682 (0.013) & \textbf{0.711 (0.023)} & 0.692 (0.004) \\
SonyAIBORobotSurface2          & 0.558 (0.018) & 0.530 (0.011) & 0.554 (0.018) & 0.569 (0.010) & \textbf{0.591 (0.033)} \\
Strawberry                     & 0.565 (0.037) & 0.612 (0.111) & 0.553 (0.038) & 0.632 (0.011) & \textbf{0.715 (0.020)} \\
TwoPatterns                    & 0.429 (0.098) & 0.494 (0.006) & \textbf{0.617 (0.174)} & 0.498 (0.008) & 0.524 (0.054) \\
Wafer                          & 0.897 (0.025) & 0.916 (0.060) & 0.885 (0.002) & 0.937 (0.030) & \textbf{0.982 (0.001)} \\
\bottomrule
\end{tabular}
\end{adjustbox}
\end{table}

\begin{figure}[!ht]
  \centering
  \includegraphics[width=\linewidth]{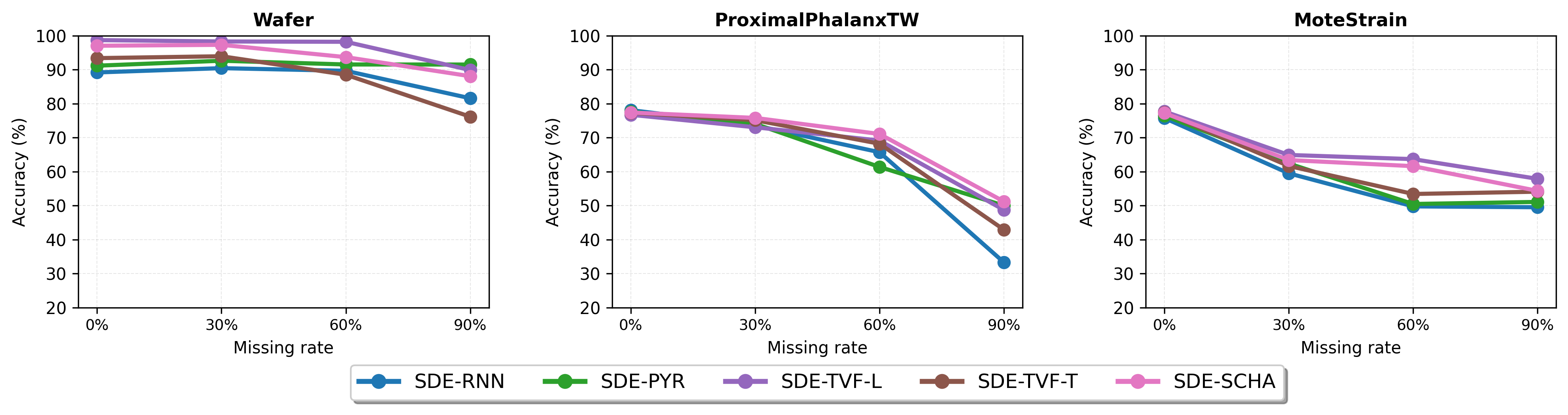}
  \caption{
  Accuracy as a function of the missing rate on three representative UCR datasets: Wafer, ProximalPhalanxTW, MoteStrain. Each curve compares the SDE--RNN baseline with four hidden level attention variants (SDE-PYR, SDE-TVF-L,
  SDE-TVF-T and SDE-SCHA).
  }
  \label{fig:ucr-missing-three}
\end{figure}

\textbf{UCR} Table ~\ref{tab:ucr-miss-60-selected-new} illustrate that even under high sparsity scenarios, SDE-TVF-L show a substantially better performance then the baseline. Concurrently, also show a slightly higher accuracy on some of the datasets than other models. 

Further, according to the Table~\ref{tab:ucr-miss-overall-means}(a), SDE-SCHA outperforms the baseline, under every missing, on overall datasets. Further, SDE-TVF-T and SDE-TVF-L are also  better than the vanilla SDE-RNN over the whole datasets. However, while the most complex model SDE-PYR often fail to improve over the baseline, might attributed to overfitting. 

As expected, Fig.~\ref{fig:ucr-missing-three} shows that performance degrades monotonically as the proportion of observed points decreases from 0\%-90\% on all models. However, the rate of degradation differs markedly across architectures. The SDE–SCHA and SDE-TVF-L consistently lie on the upper envelope of the curves or very close to it, and, more importantly, exhibits the slowest decline in accuracy as missingness increases. This effect is particularly pronounced on MoteStrain and at high missing rates on Wafer, where the baseline SDE–RNN collapses much more sharply. The pyramidal and TVF-based models generally improve over the baseline at moderate missingness but can become less stable at the most extreme sparsity levels.

\begin{table*}[!ht]
\centering
\scriptsize
\setlength{\tabcolsep}{6pt}
\renewcommand{\arraystretch}{1.1}
\caption{Average classification accuracy over the selected 10 UCR and UEA datasets under different missing rates. Best method for each missing rate is in bold.}
\label{tab:ucr-uea-miss-overall}
\begin{minipage}{0.48\linewidth}
\centering
\textbf{(a) UCR}
\vspace{0.4em}

\begin{tabular}{lcccc}
\toprule
 & \multicolumn{4}{c}{Missing rate} \\
\cmidrule(lr){2-5}
\textbf{Model} & \textbf{30\%} & \textbf{60\%} & \textbf{90\%} \\
\midrule
SDE-RNN            & 0.662 & 0.602  & 0.489  \\
SDE-SCHA           & 0.696  & 0.652  & 0.571  \\
SDE-PYR            & 0.675  & 0.617  & 0.588 \\
SDE-TVF-T          & 0.695  & 0.638  & 0.492  \\
\midrule
\textbf{SDE-TVF-L}& \textbf{0.705} & \textbf{0.679} & \textbf{0.592} \\
\bottomrule
\end{tabular}

\label{tab:ucr-miss-overall-means}
\end{minipage}
\hfill
\begin{minipage}{0.48\linewidth}
\centering
\textbf{(b) UEA}
\vspace{0.4em}

\begin{tabular}{lcccc}
\toprule
 & \multicolumn{4}{c}{Missing rate} \\
\cmidrule(lr){2-5}
\textbf{Model}  & \textbf{30\%} & \textbf{60\%} & \textbf{90\%} \\
\midrule
SDE-RNN         &\textbf{0.464}    &0.387     &0.274\\
SDE-PYR         &0.437    &0.354     &0.281\\
SDE-TVF-T       &0.452    &0.375     &0.288\\
SDE-SCHA        &0.456    &0.384     &0.307\\
\midrule
SDE-TVF-L       &0.455    &\textbf{0.411}     &\textbf{0.345}\\
\bottomrule
\end{tabular}
\label{tab:attn-miss-overall-means}
\end{minipage}
\end{table*}

\textbf{UEA} The aggregated results in Table~\ref{tab:attn-miss-overall-means} reinforce this pattern. On the multivariate UEA benchmarks, SDE-TVF-L attains the highest average accuracy for missingness levels 60\% and 90\%, reach around 2.5\% and 7\% higher, indicating that it is the most robust attention design across sparsity regimes. 

\begin{figure}[!ht]
  \centering
  \includegraphics[width=\linewidth]{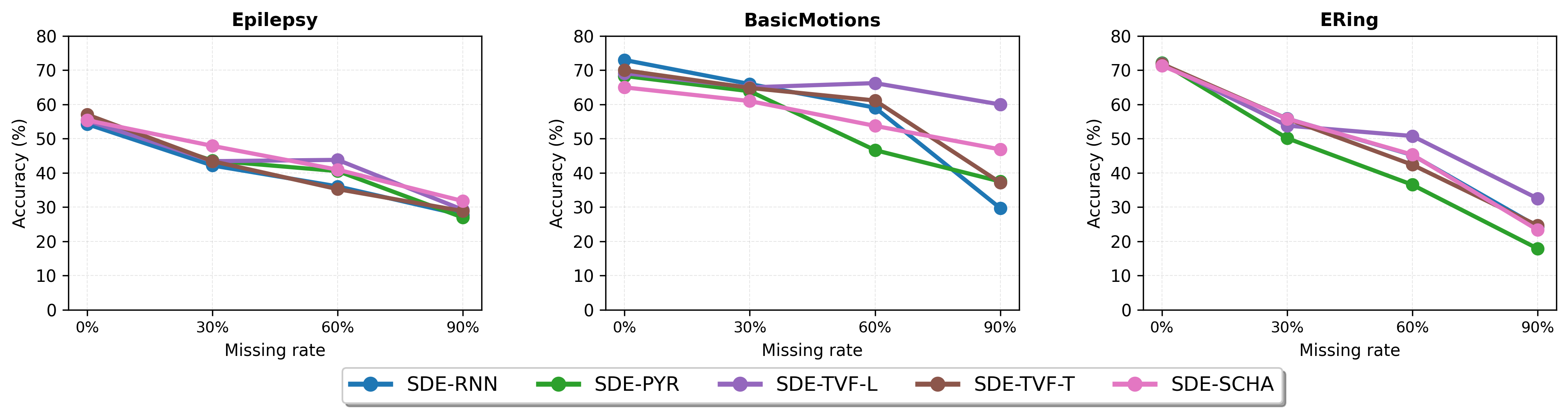}
  \caption{
  Accuracy as a function of the missing rate on three representative UEA datasets: Epilepsy, NasicMotion, Ering. Each curve compares the SDE--RNN baseline with four hidden level attention variants (SDE-PYR, SDE-TVF-L,
  SDE-TVF-T and SDE-SCHA). }
  \label{fig:uea-missing-three}
\end{figure}

In contrast to the UCR setting, Fig.~\ref{fig:uea-missing-three} shows that no single attention mechanism uniformly dominates across all datasets. The baseline SDE–RNN is often clearly suboptimal at medium–high missing rates, indicating that some form of latent attention is beneficial.

\section{Discussion}
\label{sec:discussion}

This work set out to ask whether channel-level attention can systematically improve Neural SDE–based sequence models, and which attention designs are most effective under severe observation sparsity. Overall, our results show that latent–space attention is a consistently beneficial complement to Neural SDE dynamics, and that the LSTM-based time–varying feature attention (SDE–TVF–L) offers the strongest and most stable gains among the mechanisms considered. 

On the univariate UCR datasets, all attention–augmented variants improve over the vanilla SDE–RNN in both fully observed and highly sparse regimes. The periodic toy experiment confirms that any latent–space attention reduces interpolation error relative to the backbone, with time–varying feature attention and latent channel attention providing the largest gains once a moderate fraction of points is observed. Aggregated UCR results further show that SDE–TVF–L achieves the highest mean accuracy at all missing rates, while latent channel recalibration remains a simple but reliable alternative that outperforms the baseline with negligible computational overhead. In contrast, the pyramidal multi–scale module yields only modest and less consistent improvements, suggesting that additional temporal complexity is not always necessary when the backbone already models continuous trajectories on short, low–dimensional series.

On the multivariate UEA benchmarks, the behaviour of different attention mechanisms is more heterogeneous but broadly consistent with the UCR trends. All attention modules tend to match or exceed the SDE–RNN baseline at medium to high missingness, confirming that explicit channel–level gating is helpful even for richer dynamics. When averaged across datasets and sparsity levels, SDE–TVF–L again attains the best or near–best performance, with SDE-SCHA typically ranking second. Pyramidal attention is competitive on some datasets with pronounced long–range structure but does not dominate overall, indicating that its benefits are highly dataset–dependent. In practice, these patterns suggest that SDE–TVF–L is a strong default choice when computational budget permits, whereas latent channel attention provides a cheaper, robust option in resource–constrained settings.

Our study has several limitations. We focus on a single SDE–RNN backbone with a fixed Brownian path and do not compare against strong discrete time baselines for irregular time series such as GRU–D or Transformer models with masking. Moreover, the type of distribution shift we consider is restricted to changes in the missing rate; more realistic shifts induced by different populations, sensors or acquisition protocols are not explicitly modelled. Finally, all attention modules act on the pre–RNN latent state, and we do not explore attention directly over the latent SDE trajectory or joint channel–temporal attention within the continuous dynamics. Addressing these limitations, for example, by integrating SDE–Attention into broader benchmark suites with stronger baselines, or by designing adaptive mechanisms that select or combine attention types per dataset, is a promising direction for future work, particularly in high–stakes clinical time series applications.

\section{Conclusion}
\label{sec:conclusion}

We introduced SDE--Attention, a family of attention-augmented SDE--RNN architectures for irregular and partially observed time series. Experiments on a synthetic periodic benchmark and on UCR and UEA datasets show that latent-space attention consistently improves over a vanilla SDE--RNN. Among the mechanisms considered, the LSTM-based time-varying feature attention (SDE--TVF-L) achieves the strongest and most consistent gains across missingness levels and dataset families, while latent channel attention offers a cheaper but still robust alternative. These findings suggest that explicit channel-level gating is a valuable complement to Neural SDE dynamics and offer practical guidance for choosing attention mechanisms in SDE-based sequence models.

\bibliography{main}

@inproceedings{chen2018neural,
  title        = {Neural Ordinary Differential Equations},
  author       = {Chen, Ricky T. Q. and Rubanova, Yulia and Bettencourt, Jesse and Duvenaud, David},
  booktitle    = {Advances in Neural Information Processing Systems},
  year         = {2018}
}

@inproceedings{rubanova2019latent,
  title        = {Latent Ordinary Differential Equations for Irregularly-Sampled Time Series},
  author       = {Rubanova, Yulia and Chen, Ricky T. Q. and Duvenaud, David},
  booktitle    = {Advances in Neural Information Processing Systems},
  year         = {2019}
}

@inproceedings{kidger2020neuralcde,
  title        = {Neural Controlled Differential Equations for Irregular Time Series},
  author       = {Kidger, Patrick and Morrill, James and Foster, James and Lyons, Terry},
  booktitle    = {Advances in Neural Information Processing Systems},
  year         = {2020}
}

@article{liu2019neural,
  title={Neural sde: Stabilizing neural ode networks with stochastic noise},
  author={Liu, Xuanqing and Xiao, Tesi and Si, Si and Cao, Qin and Kumar, Sanjiv and Hsieh, Cho-Jui},
  journal={arXiv preprint arXiv:1906.02355},
  year={2019}
}

@article{che2016grud,
  title        = {Recurrent Neural Networks for Multivariate Time Series with Missing Values},
  author       = {Che, Zhengping and Purushotham, Sanjay and Cho, Kyunghyun and Sontag, David and Liu, Yan},
  journal      = {Scientific Reports},
  volume       = {8},
  pages        = {6085},
  year         = {2018}
}

@misc{Seqlink2024,
      title={SeqLink: A Robust Neural-ODE Architecture for Modelling Partially Observed Time Series}, 
      author={Futoon M. Abushaqra and Hao Xue and Yongli Ren and Flora D. Salim},
      year={2024},
      eprint={2212.03560},
      archivePrefix={arXiv},
      primaryClass={cs.LG},
      url={https://arxiv.org/abs/2212.03560}, 
}

@article{rumelhart1986learning,
  title={Learning representations by back-propagating errors},
  author={Rumelhart, David E and Hinton, Geoffrey E and Williams, Ronald J},
  journal={nature},
  volume={323},
  number={6088},
  pages={533--536},
  year={1986},
  publisher={Nature Publishing Group UK London}
}

@article{graves2012long,
  title={Long short-term memory},
  author={Graves, Alex},
  journal={Supervised sequence labelling with recurrent neural networks},
  pages={37--45},
  year={2012},
  publisher={Springer}
}

@article{chung2014empirical,
  title={Empirical evaluation of gated recurrent neural networks on sequence modeling},
  author={Chung, Junyoung and Gulcehre, Caglar and Cho, KyungHyun and Bengio, Yoshua},
  journal={arXiv preprint arXiv:1412.3555},
  year={2014}
}

@article{mozer2017discrete,
  title={Discrete event, continuous time rnns},
  author={Mozer, Michael C and Kazakov, Denis and Lindsey, Robert V},
  journal={arXiv preprint arXiv:1710.04110},
  year={2017}
}

@inproceedings{shukla2018modeling,
  author    = {Satya Narayan Shukla and Benjamin M. Marlin},
  title     = {Modeling Irregularly Sampled Clinical Time Series},
  booktitle = {Proceedings of the 3rd Machine Learning for Health Workshop (ML4H) at NeurIPS},
  year      = {2018},
  archivePrefix = {arXiv},
  eprint    = {1812.00531}
}

@article{higham2001algorithmic,
  title={An algorithmic introduction to numerical simulation of stochastic differential equations},
  author={Higham, Desmond J},
  journal={SIAM review},
  volume={43},
  number={3},
  pages={525--546},
  year={2001},
  publisher={SIAM}
}

@article{oh2024stable,
  title={Stable neural stochastic differential equations in analyzing irregular time series data},
  author={Oh, YongKyung and Lim, Dong-Young and Kim, Sungil},
  journal={arXiv preprint arXiv:2402.14989},
  year={2024}
}

@article{dahale2023general,
  title={A general framework for uncertainty quantification via neural sde-rnn},
  author={Dahale, Shweta and Munikoti, Sai and Natarajan, Balasubramaniam},
  journal={arXiv preprint arXiv:2306.01189},
  year={2023}
}

@article{bagnall2018uea,
  title        = {The UEA Multivariate Time Series Classification Archive, 2018},
  author       = {Bagnall, Anthony and Dau, Hoang Anh and Lines, Jason and Flynn, Michael and Large, James and Bostrom, Aaron and Southam, Paul and Keogh, Eamonn},
  journal      = {arXiv preprint arXiv:1811.00075},
  year         = {2018}
}

@article{dau19ucr,
  title={The {UCR} time series archive},
  author={H. Dau and A. Bagnall and K. Kamgar and M. Yeh and Y. Zhu and S. Gharghabi and C. Ratanamahatana and A. Chotirat and E. Keogh},
  journal={IEEE/CAA Journal of Automatica Sinica},
  volume={6},
  number={6},
  pages={1293--1305},
  year={2019},
  publisher={IEEE}
}

@inproceedings{hu2018senet,
  title        = {Squeeze-and-Excitation Networks},
  author       = {Hu, Jie and Shen, Li and Sun, Gang},
  booktitle    = {Proceedings of the IEEE Conference on Computer Vision and Pattern Recognition},
  pages        = {7132--7141},
  year         = {2018}
}

@article{zhu2022tcran,
  title        = {TCRAN: Multivariate Time Series Classification Using Time Corrected Residual Attention Network},
  author       = {Zhu, Hong and Yang, Huaiyuan and Ding, Wei},
  journal      = {Applied Soft Computing},
  volume       = {115},
  pages        = {108117},
  year         = {2022}
}

@inproceedings{liu2022pyraformer,
  title        = {Pyraformer: Low-Complexity Pyramidal Attention for Long-Range Time Series Modeling and Forecasting},
  author       = {Liu, Shiyang and Gao, Ying and Zhang, Jing and Wang, Rui and Li, Xinbing and Yang, Zongqing},
  booktitle    = {International Conference on Learning Representations},
  year         = {2022}
}

@inproceedings{zerveas2021tstransformer,
  title        = {A Transformer-Based Framework for Multivariate Time Series Representation Learning},
  author       = {Zerveas, George and Jayaraman, Srikar and Patel, Dhaval and Bhamidipaty, Athanasios and Eickhoff, Carsten},
  booktitle    = {Proceedings of the ACM SIGKDD International Conference on Knowledge Discovery \& Data Mining},
  year         = {2021}
}
\bibliographystyle{tmlr}


\end{document}